\documentclass[10pt,twocolumn,letterpaper]{article}

\usepackage{iccv}
\usepackage{times}
\usepackage{epsfig}
\usepackage{graphicx}
\usepackage{amssymb,amsmath,amsthm}
\usepackage{booktabs}
\usepackage{url}

\newcommand{\vatex}{\textsc{VaTeX}\xspace}
\newcommand{\R}{\mathbb{R}}
\newcommand{\vfeat}{\textbf{V}^k}
\newcommand{\encs}{\textbf{E}^k}
\newcommand{\maxpool}[1]{\textsc{MaxPool}\left(#1\right)}
\newcommand{\avgpool}[1]{\textsc{AvgPool}\left(#1\right)}

\newcommand{\fflayerb}[2]{#2 \textbf{W}_{#1} + b_{#1}}

\newcommand{\tanhlayerb}[2]{\tanh\left(\fflayerb{#1}{#2}\right)}
\newcommand{\dropout}[2]{\textsc{Dropout}\left(#1, p\leftarrow #2\right)}

\usepackage[breaklinks=true,bookmarks=false]{hyperref}

\iccvfinalcopy

\ificcvfinal\fi

\begin{document}

\title{Imperial College London Submission to \vatex Video Captioning Task}

\author{Ozan Caglayan \quad Zixiu Wu \quad Pranava Madhyastha \quad Josiah Wang \quad Lucia Specia\\
Department of Computing, Imperial College London, UK\\
\small \tt \{o.caglayan, zixiu.wu18, pranava, josiah.wang, l.specia\}@imperial.ac.uk}

\maketitle
\ificcvfinal\thispagestyle{empty}\fi

\begin{abstract}
This paper describes the Imperial College London team's submission to the 2019' \vatex video captioning challenge, where we first explore two sequence-to-sequence models, namely a recurrent (GRU) model and a transformer model, which generate captions from the I3D action features. We then investigate the effect of dropping the encoder and the attention mechanism and instead conditioning the GRU decoder over two different vectorial representations: (i) a max-pooled action feature vector and (ii) the output of a multi-label classifier trained to predict visual entities from the action features. Our baselines achieved scores comparable to the official baseline. Conditioning over entity predictions performed substantially better than conditioning on the  max-pooled feature vector, and only marginally worse than the GRU-based sequence-to-sequence baseline.
\end{abstract}

\section{Introduction}
Video captioning is a challenging task at the intersection of spatio-temporal visual processing and natural language processing.
The current state of the art in video captioning is dominated by end-to-end deep neural networks which generate natural language sentences based on a combination of object, action and optical flow features extracted from input videos \cite{Aafaq2018}. The task becomes even more interesting when videos are captioned in a multilingual fashion, as in the case of the recent \vatex dataset which provides English and Chinese captions for a large collection of 35,000 ``open domain'' videos \cite{wang2019vatex}. A video captioning challenge\footnote{\url{http://vatex.org/main/captioning.html}} has been organized around the dataset.
This work presents our efforts for the English video captioning sub-task. We explore end-to-end sequence-to-sequence models \cite{Bahdanau2014, Sutskever2014, Venugopalan2015} and two-stage neural approaches to first detect entities from the videos and then map the set of entities onto natural language sentences. We hypothesize that distilling semantic entities from visual features will reveal a stronger and more structured input representation than encoding the latent video features directly. To achieve this, we train a multi-label entity tagger \cite{Fang_2015_CVPR, You_2016_CVPR} to predict the nouns, verbs and adjectives from the spatio-temporal I3D action features \cite{I3D} the challenge organizers provided. The label set for the tagger is constructed from the training set captions using a part-of-speech tagger. We then use these weak entity labels in different ways to generate the English captions and compare the performance to the recurrent \cite{Bahdanau2014} and transformer-based \cite{transformer} baselines.

Our primary findings are as follows: (i) both sequence-to-sequence baselines perform similarly to the official challenge baseline \cite{wang2019vatex}, with the transformer-based model obtaining marginally better BLEU and CIDEr scores; (ii) simply initializing the hidden state of a GRU language model with entity prediction scores is only 0.7 BLEU inferior to the recurrent sequence-to-sequence model; (iii) replacing the entity score vector with the max-pooled I3D action feature vector resulted in a substantial performance loss. The latter point potentially indicates that a recurrent decoder is not able to fully exploit the latent information encoded in the action features.

\section{Dataset}
We train our systems on the \vatex dataset~\cite{wang2019vatex} which provides 25,991 training, 3,000 validation and 6,000 test videos with 10 English and 10 Chinese captions each. For video representation, we rely on the provided spatio-temporal action features extracted from a pre-trained I3D ConvNet~\cite{I3D}. For a given video $k$, the associated video representation is a sequence of 1024-dimensional feature vectors, which will be denoted by $\vfeat \in \R^{N\times 1024}$.

Focusing on the English captions only, we apply a standard pre-processing pipeline where each English caption is punctuation-normalized, lowercased and tokenized using Moses scripts~\cite{Moses:2007:acl}. The final vocabulary considers words that occur at least 5 times in the training set and has a size of 10,300 tokens. For the transformer-based model only, we used subword segmentation using the BPE algorithm \cite{sennrich2015neural} with the number of merge operations set to 16,000.

\section{Methods}
\subsection{Baselines}
\subsubsection{GRU-based S2S Model}
\label{sec:grus2s}
Our recurrent sequence-to-sequence model consists of a video encoder followed by an attentive decoder \cite{nematus}. Namely, the encoder is a 2-layer bi-directional GRU \cite{cho2014gru} which receives as input the linear projection of $\vfeat$ and produces the sequence of hidden states $\encs$ as follows:
\begin{align*}
    \textbf{F}^k &= \fflayerb{v}{\vfeat}\\
    \textbf{H}^k &= \textsc{Encoder}\left(\textbf{F}^k, h_{0}^k\leftarrow 0\right)\\\label{eq:enc}
    \encs &= \dropout{(\fflayerb{e}{\textbf{H}^k})}{0.4}
\end{align*}

The decoder consists of two GRU layers as well with an attention block \cite{Bahdanau2014} in the middle. At each timestep, the decoder computes the conditional probability of the next token based on the previous token history and the video representation. For the sake of clarity, the sample index $k$ is omitted in the following equations:
\begin{align*}
    h'_{t} &= \textsc{GRU}_1\left(y_{t-1}, h''_{t-1}\right)\\
    z_{t} &= \textsc{MLP-Attention}\left(\textbf{E}, h'_{t}\right)\\
    h''_{t} &= \textsc{GRU}_2\left(z_t, h'_{t}\right)\\
    o_t &= \dropout{\tanhlayerb{o}{h''_{t}}}{0.5}\\
    P\left(y_t | y_{<t}, \textbf{E}\right) &= \textsc{SoftMax}\left(\fflayerb{s}{o_t}\right)
\end{align*}
At the beginning of the decoding, the hidden state of the first GRU is initialized with the non-linear transformation of the average-pooled encoding $\avgpool{\textbf{E}^k}$.

\subsubsection{Transformer-based S2S Model}
Our transformer-based \cite{transformer} model consists of a 6-layer encoder and a 6-layer decoder, with hidden size $d$ and the number of heads set to 1024 and 16, respectively. The input to the encoder is the sequence of feature vectors $\vfeat$. The rest of the architecture functions in the same way as the canonical text-based transformer.

\subsection{I3D-Conditioned GRU Decoder}
\label{sec:veccond}
In this model, we first replace the bi-directional encoder of our GRU baseline (Section~\ref{sec:grus2s}) with a linear layer followed by a max-pooling operator. This encodes the video into a single vector video representation $v$. We then remove the attention and the second GRU from the decoder and condition the remaining GRU over $v$ through its initial state $h_{0}$:
\begin{align*}
    v^k       &= \maxpool{\fflayerb{v}{\vfeat}}\\
    h_{0}^k &= \tanhlayerb{r}{v^k}
\end{align*}

\subsection{Going from Entities to Captions}
Instead of going end-to-end from action features to captions, here we design a two-stage paradigm where we first train a video tagger to predict entities from the provided action features. We conjecture that discretizing the video representation into a set of entities is a form of denoising which could be helpful in obtaining better captions.

In order to extract the labels for the videos, we feed the detokenized English training captions to Stanford Part-of-Speech (POS) tagger~\cite{toutanova-etal-2003-feature} and obtain\footnote{We use the \texttt{english-caseless-left3words-distsim} model for tagging.} POS tags along with the lemmatized tokens for each caption. We only keep lemmas with POS tags matching the \texttt{(NN*|VB*|JJ*)} pattern \ie nouns, verbs and adjectives. The set of entities for a given video is then obtained by taking the union of lemmas collected from each of the 10 captions associated to the video. The final entity vocabulary for the training set consists of \textbf{4,383} tokens that occur at least 10 times across the training set entities.

\paragraph{Video Tagger.}
\begin{figure}[t]
\centering
\includegraphics[width=.47\textwidth]{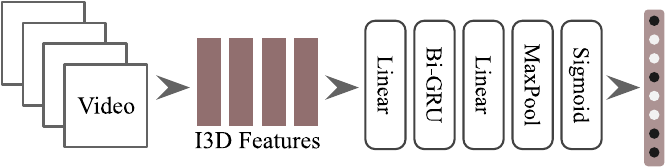}
\caption{Video tagger architecture: the output is a multi-label prediction vector of size 4,383 \ie the size of the entity vocabulary.}
\label{fig:tagger}
\end{figure}

We experiment with various classifier architectures and select the one shown in Figure~\ref{fig:tagger} based on hyperparameter and architecture search. We use a multi-label objective function which considers each prediction of the sigmoid output layer as a one-vs-all binary classification problem. Specifically, for a given video $k$, the loss $\textsc{L}(k)$ is defined as:
\begin{align*}
    p^{k} &= \textsc{Tagger}(\vfeat) \quad\quad p^{k} \in \R^{4383} \\
    \textsc{L}(k) &= -\frac{1}{C}\sum_c{w_c\left(y^{k}_c\log\left(p^{k}_c\right) + (1 - y^{k}_c)\log\left(1 - p^{k}_c\right)\right)}
\end{align*}
To remedy label imbalance, we scale each term in the loss formulation with $w_c$ during training to penalize terms related to rare entities more than the frequent ones ($\times10$ for the least frequent class). The model is early-stopped based on F1 score with classifier threshold set to $0.3$.
Being relatively small with only 0.9M parameters, the final model obtains an F1 score of 37.7 on validation set.

\subsubsection{Entity-Conditioned GRU Decoder}
This is a modification of the I3D-conditioned decoder (Section~\ref{sec:veccond}) where we simply replace the pooled visual vector $v^k$ with the entity prediction vector $p^k \in \R^{4383}$ extracted from the pre-trained video tagger. This model is important in the sense that it may reveal whether entity distillation from action features conveys a better signal than the original action representations or not.

\subsubsection{Top-$k$ Attention}
This model modifies the baseline GRU from Section~\ref{sec:grus2s} and replaces the input feature sequence with top-$k$ entity embeddings. In fact, this is exactly a neural machine translation (NMT) system where the input is a set of entities instead of source sentences. 
Note that the embeddings are trained from scratch \ie they are not initialized with external pre-trained representations. We experiment with $k \in \{5, 10, 50, 100\}$. For example, for $k\mathrm{=}5$, a video from the validation set is tagged with \textit{\{perform, dancer, costume, dance, dress\}} with one of the ground truth captions being: \textit{``a woman in costume belly dancing on stage"}.


\section{Results}
\begin{table}[t]
\centering
\resizebox{.47\textwidth}{!}{%
\renewcommand*{\arraystretch}{1.1}
\begin{tabular}{lccc}
\toprule
                                & BLEU      & METEOR    & CIDEr     \\ \midrule
I3D Conditioning                & 27.5      & 21.4      & 4.48      \\
Entity Conditioning\phantom{xxxxx} & 29.8      & 22.5      & 5.34      \\ 
\bottomrule
\end{tabular}%
}
\vspace*{.3em}
\caption{\textbf{Validation set} scores for decoder conditioning.}
\label{tbl:conditioned}
\end{table}

\begin{table}[t]
\centering
\resizebox{.47\textwidth}{!}{%
\renewcommand*{\arraystretch}{1.1}
\begin{tabular}{lccc}
\toprule
                                        & BLEU      & METEOR    & CIDEr     \\ \midrule
$k = 5$                                 & 25.8  & 20.5      & 3.97  \\
$k = 10$                                & 27.9  & 21.7      & 4.80  \\
$k = 100$                               & 28.8  & 22.3      & 5.23  \\
$k = 50$                                & 29.6  & 22.3      & 5.29  \\ \midrule
Entity Conditioning\phantom{xxxxx}     & 29.8      & 22.5      & 5.34      \\
\bottomrule
\end{tabular}%
}
\vspace*{.3em}
\caption{\textbf{Validation set} scores for Top-$k$ attention.}
\label{tbl:topk}
\end{table}
We first compare the proposed models based on the validation set scores and then select a subset of the models and report public test set results through the challenge's leaderboard. The BLEU \cite{Papineni:2002:acl}, METEOR \cite{meteor} and CIDEr \cite{vedantam2015cider} scores for the validation set are computed using the \texttt{coco-caption} toolkit\footnote{\url{https://github.com/tylin/coco-caption}}.

Table~\ref{tbl:conditioned} compares the I3D and entity-conditioned GRU decoders. We see that conditioning over the entity prediction vector performs substantially better than conditioning over the max-pooled I3D action feature vector. This seems to suggest that going explicitly from latent action features to entity space provides a richer representation for the task at hand.
As to the Top-$k$ attention experiments, we observe that a set of top $\sim$50 entities is as expressive as conditioning on top of the full entity detection scores (Table~\ref{tbl:topk}).


\begin{table}[t]
\centering
\resizebox{.47\textwidth}{!}{%
\renewcommand*{\arraystretch}{1.1}
\begin{tabular}{lccc}
\toprule
                                & BLEU      & METEOR    & CIDEr     \\ \midrule
Top-$50$ Attention              & 26.2      & 20.5      & 3.91      \\
Entity Conditioning             & 27.0      & 20.8      & 4.04      \\ \midrule
GRU-S2S                         & 27.7      & 21.6      & 4.43      \\
\vatex-S2S \cite{wang2019vatex} & 28.5      & 21.6      & 4.51      \\
Transformer-S2S$^\dagger$\phantom{xxxxxx}       & 29.0      & 21.1      & 4.52      \\
\bottomrule
\end{tabular}%
}
\vspace*{.3em}
\caption{Overall \textbf{test set} results: $\dagger$ marks our submission.}
\label{tbl:baselines}
\end{table}

Finally, Table~\ref{tbl:baselines} shows the public leaderboard performance of our systems with respect to the official recurrent baseline \cite{wang2019vatex}. Our sequence-to-sequence systems and the \vatex baseline
are in the same ballpark, with the transformer model obtaining the best BLEU and CIDEr score\footnote{The difference is unlikely to be due to word-based and BPE-based vocabulary difference across the baselines since at test time, none of the tokens generated by the BPE-based transformer were subwords.}. It should be noted that the transformer model is much deeper than the recurrent ones in terms of the number of encoder and decoder layers.
The entity-conditioning model lags behind the GRU-S2S by only 0.7 BLEU, demonstrating that a sequence-to-sequence model does not perform dramatically better than a simple bag-of-entities system on this dataset.

\section{Conclusion}
This paper summarizes our English video captioning efforts for the \vatex challenge. We first start by adapting the recurrent and transformer sequence-to-sequence frameworks in order to perform caption generation based on a sequence of action features. We contrast the performance of these widely-known baselines to conditional language models where we simply initialize the decoder either with an aggregate action feature vector obtained through max-pooling or the output of a multi-label classifier trained to predict visual entities from the action features. We show that the latter performs substantially better than the former, and only marginally worse than a much more sophisticated GRU-based sequence-to-sequence baseline. This indicates that in some cases learning a mapping from action-oriented features to visual entities may provide a more expressive signal for captioning, compared to the raw features themselves.



\section*{Acknowledgments}
This work was supported by the MultiMT (H2020 ERC Starting Grant No. 678017) and MMVC (Newton Fund Institutional Links Grant, ID 352343575) projects.

{\small
\bibliographystyle{ieee_fullname}
\bibliography{refs}
}

\end{document}